\documentclass[double]{article}
\usepackage{times}
\usepackage{algorithm}
\usepackage{algorithmic}
\usepackage{wrapfig}
\usepackage{verbatim}
\usepackage{amsmath}
\usepackage{graphicx}
\usepackage{subcaption}
\usepackage[active]{srcltx}
\usepackage{color}
\usepackage{lineno}
\usepackage{dirtytalk}
\usepackage{amsthm}
\usepackage{authblk}
\usepackage{listings}

\usepackage{epsfig,graphicx,amsfonts}
\usepackage{amsmath,amsthm,amssymb}
\usepackage{xcolor}

\usepackage{adjustbox}
\usepackage{multirow}
\usepackage{setspace}
\usepackage{hyperref}
\usepackage{comment}

\long\def\comment #1\commentend{}

\begin{document}

\title{\Large Mathematical Modeling of BCG-based Bladder Cancer Treatment Using Socio-Demographics}
\author{Elizaveta Savchenko$^{1*}$, Ariel Rosenfeld$^{2}$, Svetlana Bunimovich-Mendrazitsky$^{1}$\\
\(^1\) Department of Mathematics, Ariel University, Ariel, Israel\\
\(^2\)  Department of Information Science, Bar Ilan University, Ramat-Gan, Israel\\
\(*\) Corresponding author: svetlanabu@ariel.ac.il

}

\date{}

\maketitle 

\begin{abstract}
Cancer is one of the most widespread diseases around the world with millions of new patients each year. Bladder cancer is one of the most prevalent types of cancer affecting all individuals alike with no obvious \say{prototypical patient}. The current standard treatment for BC follows a routine weekly Bacillus Calmette–Guérin (BCG) immunotherapy-based therapy protocol which is applied to all patients alike. The clinical outcomes associated with BCG treatment vary significantly among patients due to the biological and clinical complexity of the interaction between the immune system, treatments, and cancer cells. In this study, we take advantage of the patient's socio-demographics to offer a personalized mathematical model that describes the clinical dynamics associated with BCG-based treatment. To this end, we adopt a well-established BCG treatment model and integrate a machine learning component to temporally adjust and reconfigure key parameters within the model thus promoting its personalization. Using real clinical data, we show that our personalized model favorably compares with the original one in predicting the number of cancer cells at the end of the treatment, with \(14.8\%\) improvement, on average. \\ \\

\end{abstract}

\maketitle \thispagestyle{empty}

\pagestyle{myheadings} \markboth{Draft:  \today}{Draft:  \today}
\setcounter{page}{1}

\section{Introduction}
\label{sec:introduction}
Cancer is one of the most widespread illnesses in the world, responsible for millions of death every year with increasing numbers over time \cite{open_intro}. Bladder Cancer (BC) is the seventh most common cancer worldwide, associated with 400 thousand new cases and 150 thousand deaths every year as of 2018 \cite{Jemal} and 600 thousand yearly new cases worldwide with only 77\% five-year survival rate as of 2022\footnote{We refer the interested reader to the full updated statistics at \url{https://www.cancer.net}}. BC has many forms and clinical stages, mainly differing by the depth of the cancer cell population in the urothelium \cite{bc_numbers}. In the scope of this study, we focus on the non-invasive (superficial) BC where the cancer cells do not spread beyond the inner layer of the bladder where the entire cancer cell population is located inside the urothelium and does not invade other tissues. The non-invasive BC is highly common with roughly four out of five of all BC cases being diagnosed at the non-invasive stage \cite{noninvasive_bc}. In these cases, multiple treatment protocols exist including chemotherapy-based \cite{bc_chemotherapy} and immunotherapy-based \cite{bcg_best} treatments. Currently, the immunotherapy treatment suggested by \cite{MoralesEidingerBruce} that follows weekly injections of Bacillus Calmette–G\'{e}rin (BCG) seems to achieve the best clinical improvement over a broad spectrum of clinical states  \cite{bcg_good1,bcg_good3}. Most notably, BCG-based immunotherapy treatment has proven to reduce both the recurrence and progression of BC \cite{bcg_good2}. The BCG treatment protocol is defined by the amount of the injected dosage, the number of injections, and the schedule of the treatment \cite{Guzev}. Any change in one or more of these configurations can have a drastic effect on the patient's clinical state. However, due to the complexity of the biological dynamics, it is challenging to predict this change in advance.

\begin{figure}
\centering
\includegraphics[width=0.75\textwidth]{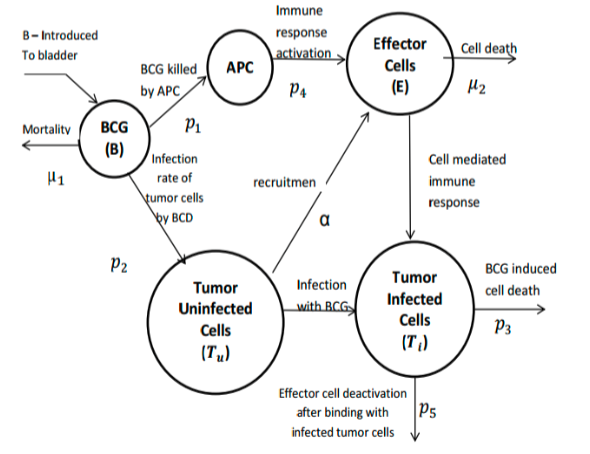}
\caption{Cell population dynamics in the bladder - taken with permission from \cite{BunimovichShochat}. BCG (\(B\)) stimulates effector cells (\(E\)) of the immune system via APC activation. In addition, BCG infects uninfected cancer cells (\(T_u\)) which recruit effector cells into the bladder. Infected cancer cells (\(T_i\)) are destroyed by effector cells.}
\label{fig:figure_1}
\end{figure}

In order to derive a suitable treatment protocol for patients, either at the individual or group level, researchers and clinicians often leverage the power of mathematical models and simulation \cite{CastiglionePiccoli}. Commonly, \textit{in silico} experiments provide a cheap, quick, and humane solution for clinical investigation of treatment protocols, allowing one to better understand and capture the underlying pharmacokinetics and pharmacodynamics \cite{Byrne,teddy_nano}. These models and simulations typically rely on an ordinary differential equation (ODE) representation where each variable describes a different cell population size \cite{Kuznetsov,KimSteinberg,KirschnerPanetta,Panetta,PillisRadunskaya}. 
Indeed, the modeling and simulation of BC treatment protocols using this approach have attracted much attention in the literature \cite{agent_spatial_location,BunimovichGoltser,Nave}. Notably, \cite{BunimovichShochat} proposed a BCG-based treatment protocol for BC which assumed continuous BCG instillation with a logistic growth for cancer cells inside the bladder. Formally, the proposed model takes the form:

\begin{equation}
\frac{dB(t)}{dt} = -p_1E(t)B(t) - p_2B(t)T_u (t) - \mu_1 B(t) + b 
\end{equation}
\begin{equation}
\frac{dE(t)}{dt} = -\mu_2E(t) + \alpha T_i (t) + p_4 E(t)B(t) - p_5(t) E(t) T_i (t)
\end{equation}
\begin{equation}
\frac{dT_i (t)}{dt} = p_2 B(t)T_u (t) - p_3 T_i (t) E(t)
\end{equation}
\begin{equation}
\frac{dT_u (t)}{dt} = \lambda(t) T_u (t) - p_2 B(t)T_u (t),
\end{equation}
where \(B(t)\), \(E(t)\), \(T_i (t)\), and \(T_u (t)\) represent the concentration of BCG in the bladder, effector cell population size, the population of cancer cell that has been infected with BCG size, and the population of cancer cell that is uninfected with BCG size, respectively. The model's parameters represent the following quantities: \(p_1\) is the rate of BCG killed by effector cells; \(p_2\) is the infection rate of uninfected cancer cells by BCG; \(p_3\) is the rate of destruction of cancer cell infected by BCG by effector cells; \(p_4\) is the immune response activation rate; \(p_5\) is the rate of effector cells deactivation after binding with infected cancer cells. \(\alpha \) is the growth rate of effector cell population; \(\lambda \)  is the cancer's population growing rate; \(b\) is the amount of BCG injected to the bladder. Fig. (\ref{fig:figure_1}) shows a schematic view of this model, including the different cell populations and the interactions between them. 

One promising avenue for improving BCG-based treatment protocols is changing the currently practiced \say{one-size-fits-all} approach with a more personalized one \cite{math_good_1,teddy_cells}. Specifically, by taking social and behavioral factors into account, one is likely to obtain a more favorable treatment prediction model at the individual patient's level and lead to better clinical outcomes as treatment optimization models would be able to have an underlined more accurate outcome prediction model \cite{bcg_need_change_2}. A patient's social and behavioral characteristics can be typically extracted with minimal overhead (e.g., by a simple questioning or directly from the patient's electronic health record) as opposed to alternative information-gathering efforts such as additional clinical tests which are associated with substantial operational costs. Following this line of thought, in this work, we propose a novel mathematical model which significantly extends that of \cite{BunimovichShochat}. Our novelty lies primarily in the integration of a machine learning component which is used to assess and adjust the model's parameters over individuals and over time. Using real-world data of \(N=417\) patients, we show that our model favorably compares to the existing models.

The rest of the paper is organized as follows: Section \ref{sec:model} formally presents the framework, followed by Section \ref{sec:theory} which provides theoretical outcomes regarding the proposed model. Next, Section \ref{sec:results} outlines the results of using the proposed model on real-world clinical data. Finally, in section \ref{sec:discussion}, we analyze and discuss the results as well as propose possible future work directions.  

\section{Mathematical Modeling}
\label{sec:model}
\begin{figure}[!ht]
    \centering
\includegraphics[width=0.75\textwidth]{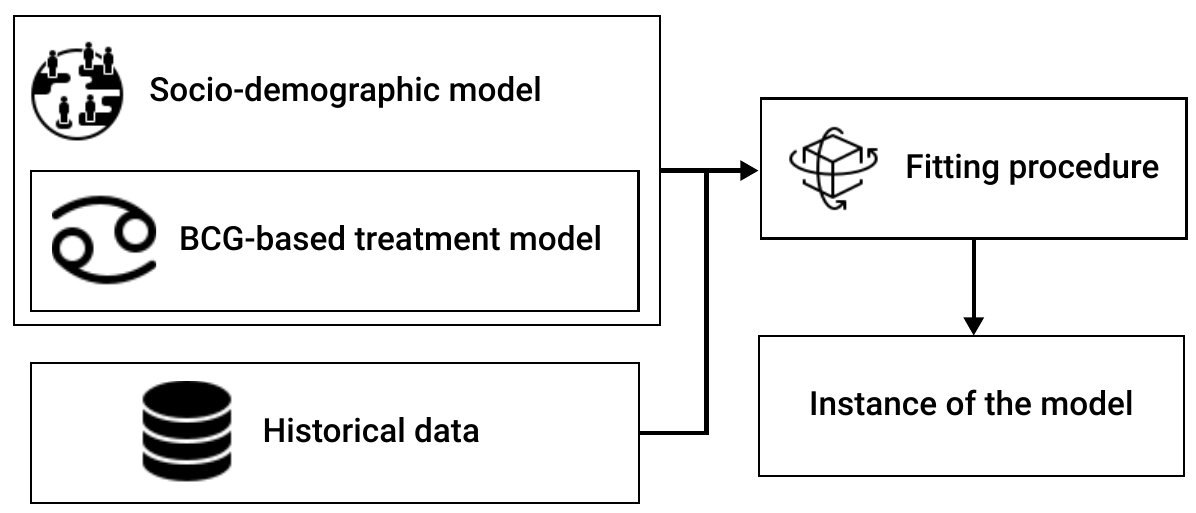}
    \caption{A schematic view of the mathematical modeling.}
    \label{fig:method}
\end{figure}

Our model consists of two interconnected modules: a BCG-based treatment module and a socio-demographic personalization module. First, we define the bio-clinical dynamics of BCG treatment for BC. Then, we formalize the socio-demographics that underline the BCG-treatment dynamics. Based on these two modules, we formulate a fitting procedure to set the parameters of an instance of the framework using historical data. Fig. \ref{fig:method} shows a schematic view of the mathematical modeling.

\subsection{BCG-based treatment module}
Our following mathematical formulation relies on extensive prior literature which proposed and analyzed several biological models to describe the biological process underlying the BCG-based immunotherapy treatment for BC with increasing levels of complexity, capturing biological and clinical properties with great levels of detail and, presumably, accuracy \cite{teddy_extended,bc_models_3,bc_models_2,bc_ode_model_3,bc_models_1,agent_spatial_location}. These and similar models describe, in a mathematical manner, the change in several cell populations over time due to (spatio-)temporal interaction between these cell populations \cite{pde_study_2,newPaper,pde_study_1}. 
Generally speaking, the main line of work for modeling BCG-based treatment for BC, which we also follow in this work, was proposed by \cite{BunimovichGoltser}. The authors used a system of ODEs that represents the cell population sizes of several cell types over time. In particular, they divide the cell population into three main groups: BCG-infected, cancer, and immune-related cells, and described their interaction. In addition, special attention was placed on the distinction between BCG-infected and non-BCG-infected cells. 

Here, we extend the model proposed by \cite{BunimovichShochat} in three manners:
First, we replace the continuous BCG injection which assumes BCG is injected at some rate at any point in time with a discrete one which assumes a set of points in time in which BCG is injected. The latter more closely describes how BCG administration is provided in practice \cite{BunimovichGoltser}; Second, we consider the uninfected cancer cell elimination by immune system cells \cite{oldPaperOde}; Third, we introduce a healthy cell population and its interactions with the other cell types during a BCG treatment \cite{teddy_cells}. Importantly, following \cite{bio_bg}'s work, we assume that the BCG interaction with healthy and cancer cells is different following the difference between the cells' surfaces as well as shape. 
In addition, in order to allow personalization within the model, we replace the scalar parameters with functions that depend on time and the socio-demographics of the patient and add a term to the immune cells population that is associated with the presence of immune cells based on the level of the patient's activeness. Hence, the model takes the form (and explained right after):

\begin{equation}
\frac{dB(t)}{dt} = \sum_{m=0}^{N-1}b\delta(t-m\tau) - p_1(t)E(t)B(t) - p_2(t)B(t)T_u (t)  - p_8(t)B(t)H_u (t) - \mu_B B(t).
\label{eq:1}
\end{equation}
\begin{equation}
\frac{dE(t)}{dt} = -\mu_E(t) E(t) + \alpha \big ( T_i (t) +  H_i (t) \big ) + p_4(t) E(t)B(t) - p_5(t) E(t) T_i (t) - p_6(t) E(t) H_i (t).
\label{eq:2}
\end{equation}
\begin{equation}
\frac{dT_i (t)}{dt} = p_2(t) B(t)T_u (t) - p_3(t) T_i (t) E(t).
\label{eq:3}
\end{equation}
\begin{equation}
\frac{dT_u (t)}{dt} = \lambda(t) T_u (t) - p_2(t) B(t)T_u (t) -p_3(t)T_u(t)E(t).
\label{eq:4}
\end{equation}
\begin{equation}
\frac{dH_u (t)}{dt} = p_7(t)H_u(t) \big ( 1 - \frac{H_u(t) + H_i(t) + T_u(t) + T_i(t)}{H_m} \big ) - p_8(t) B(t) H_u(t).
\label{eq:5}
\end{equation}
\begin{equation}
\frac{dH_i (t)}{dt} = p_8(t) B(t) H_u(t) - p_9(t)E(t)H_i(t).
\label{eq:6}
\end{equation}

In Eq. (\ref{eq:1}), \(\frac{dB(t)}{dt}\) is the dynamic change of BCG in the bladder over time. It is affected by the following five terms. First, a quantity \(b\) of BCG has been instilled into the bladder every \( \tau \) steps in time. As the instillation of the BCG is modeled by a shifted Dirac delta function \(\delta(t – m \tau), m \in \{0, \dots, N – 1\}\), the \(m_{th}\) dose raises \(B(t)\) by \(b\) units at time \(t = mr\). Second,  BCG is eliminated by the immune cells (APCs) at a rate \(p_1(t)\). Third and Fourth, BCG penetrates into uninfected cancer and uninfected healthy cells and is removed from the volume of the bladder while converting these cells into BCG-infected cancer cells at rates \(p_2(t)\) and \(p_8(t)\), respectively. Finally, the BCG cell population naturally decays at a rate \( \mu_B \).

In Eq. (\ref{eq:2}), \(\frac{dE(t)}{dt}\) is the dynamic number of immune cells over time. It is affected by the following five terms. First, the immune cell population naturally decays at a rate \( \mu_E(t) \). Second, immune cells are recruited due to the detection of BCG-infected cancer and healthy cells at a rate \(\alpha(t)\). Third, immune cells are recruited due to bacterial infection in the bladder at a rate \(p_4(t)\). Fourth and fifth, immune cells are destroyed while eliminating BCG-infected cancer and regular cells at rates \(p_5\) and \(p_6\), respectively.

In Eq. (\ref{eq:3}), \(\frac{dT_i(t)}{dt}\) is the dynamic number of BCG-infected cancer cells over time. It is affected by the following two terms. First, BCG-infected cancer cells generated from uninfected cancer cells that interacted with BCG at a rate \(p_2(t)\). Second, BCG-infected cancer cells are eliminated by immune cells at a rate \(p_3(t)\).

In Eq. (\ref{eq:4}), \(\frac{dT_u(t)}{dt}\) is the dynamic number of uninfected cancer cells over time. It is affected by the following three terms. First, uninfected cancer cells naturally grow at a rate \(\lambda\). Second, uninfected cancer cells become BCG-infected and are eliminated by immune cells at a rate \(p_2(t)\). Third, immune system cells eliminate uninfected cancer cells at a rate \(p_3\).

In Eq. (\ref{eq:5}), \(\frac{dH_u(t)}{dt}\) is the dynamic number of uninfected healthy cells over time. It is affected by the following two terms. First, healthy cells are generated to fulfill the volume of the bladder, \(H_m\), at a rate \(p_7(t)\). Second, uninfected healthy cells become BCG-infected due to the presence of BCG at a rate \(p_8(t)\).

In Eq. (\ref{eq:6}), \(\frac{dH_i(t)}{dt}\) is the dynamic number of BCG-infected healthy cells over time. It is affected by the following two terms. First, uninfected healthy cells generated from uninfected cancer cells that interacted with BCG at a rate \(p_8(t)\).  Second, BCG-infected healthy cells are eliminated by immune cells at a rate \(p_9(t)\).

         
         
         


For the proposed model, the initial condition at the beginning of the BCG treatment takes the form:
\begin{equation}
    B(0) = 0, E(0) = e > 0, T_i(0) = 0, T_u(0) = T_0 > 0, H_u(0) = H_m - T_0, H_i(0) = 0.
\end{equation}

A theoretical analysis of the model, proving that it is well-posed and analyzing its equilibria states and their stability, is provided in Section \ref{sec:theory}.

\subsection{Socio-economic parameters}
Patients are categorized into one of \(72\) socio-demographic groups. These groups are constructed based on the Cartesian product of four discretized properties: age (19-25, 26-35, 36-45, 46-55, 56-65, and 66+), gender (male and female), smocking behavior (smoker and non-smoker), and weight (underweight, normal weight, and over-weight) such that the groups are pairwise disjoint. We decided on this democratization as it is commonly used in clinical studies and would be utilized later in this study as well \cite{final_2_4,final_2_3,duration_1,test_accuracy}. That said, the model is agnostic to the democratization of these parameters. Formally, each patient is represented by a timed finite state machine \cite{fsm} as follows: \(p := (a, g, s, w)\) where \(a\) is the age group, \(g\) is the gender group, \(s\) is the smoking group, and \(w\) is the weight group. Importantly, it is assumed that a patient's socio-demographic properties do not change over the treatment process.

\subsection{Treatment model fitting}
\label{sec:fitting}
Any model is as accurate as its fitting procedure allows. Thus, one may need to fit the proposed model on historical clinical data, obtaining a good approximation of the model's parameters' values. For the proposed model and in the clinical context it occurs, the historical data commonly have several limitations that make it challenging in the best case and infeasible in the worst case to use. Specifically, the BCG treatment data generally focuses on the treatment's clinical outcome and rarely, if any, include the cell population sizes (\(E, T_i, T_u, H_i,\) and \(H_u\)) during the time of the treatment. As such, only two sample points during the course of the treatment are typically available - one at the beginning of the treatment and another one at its end. Second, the socio-demographic data is not directly taken into consideration in the model. However, it is correlated with the model's parameters. Third, the amount of available data is relatively small (usually, several hundred samples). Following the first point, and since measuring the amount of cancer cells is both clinically challenging and expensive, we assume that the data takes the form of \(T_i(0) + T_u(0), T_i(t_f) + T_u(t_f), \zeta\) where \(t_f \in \mathbb{N}\) is the time at the end of the treatment and \(\zeta \in \mathbb{R}^x\) is a vector of the socio-demographic properties of the patient such that \(x \in \mathbb{N}\) is the number of socio-demographic properties. 

\begin{figure}[!ht]
    \centering
    \includegraphics[width=0.99\textwidth]{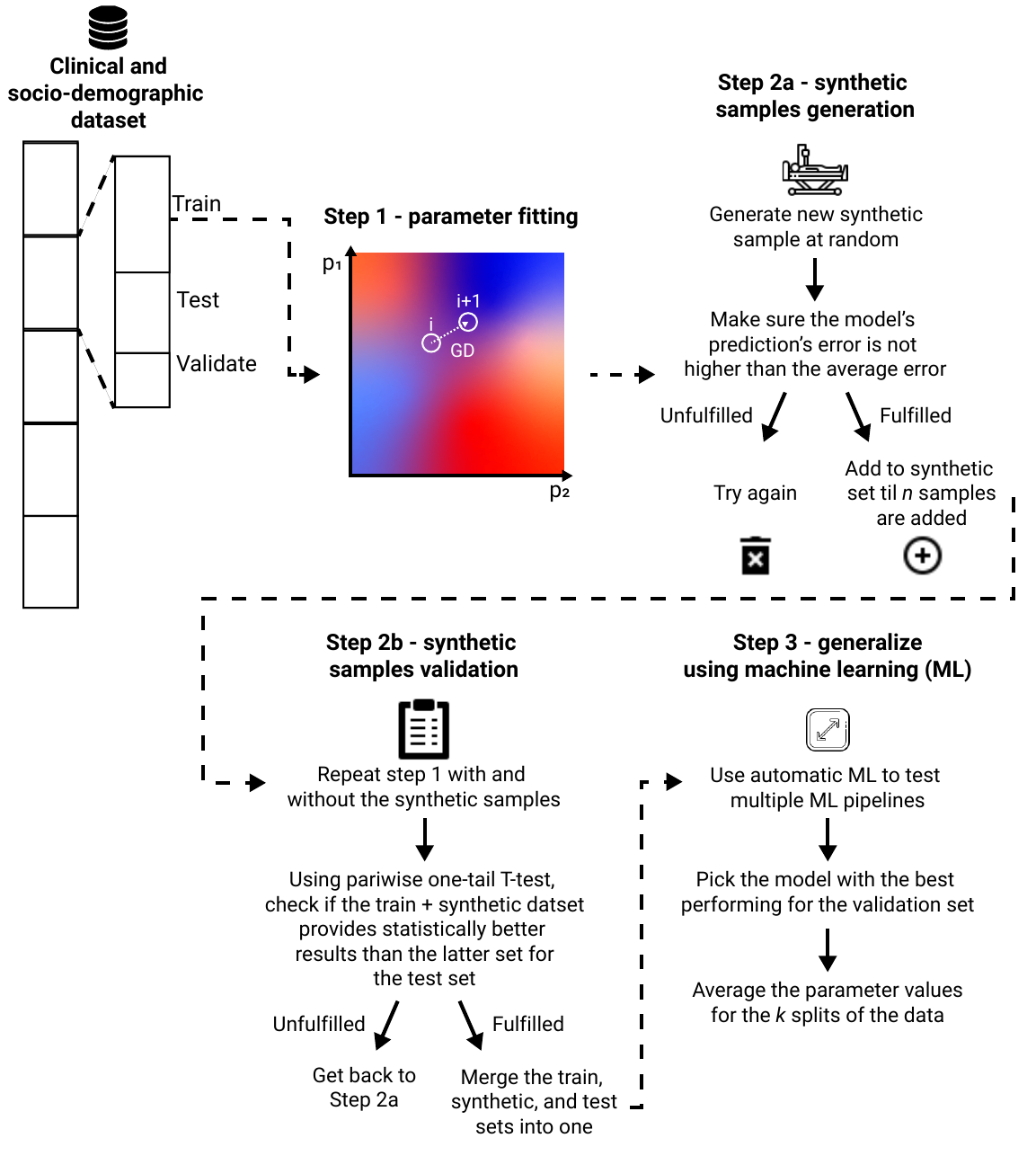}
    \caption{A schematic view of the proposed fitting procedure.}
    \label{fig:fit}
\end{figure}

During the fitting procedure, one is required to work with only the beginning and end point of the model which themselves are only providing partial knowledge of the model's state (i.e., \(T_u\) and \(T_i\) without \(B, E, H_u, H_m\)). Hence, traditional fitting procedures might obtain unrealistic courses between these two points as long as the model closely crosses near them. To tackle this challenge, we proposed a three-step fitting procedure where each step is responsible to improve the accuracy and robustness of the model. First, we divide the data into \(k\) cohorts following the \(k\)-fold cross-validation method \cite{k_fold}. This step is common in machine learning practices and improves the procedure's robustness \cite{k_fold}. Then, each cohort is further divided into train, test, and validate sets. Using only the train set, we perform the first step. Namely, we adopt the fitting procedure proposed in \cite{liza_latest} which, given the model's initial condition, the parameter space, historical data, and a loss function \(d\), we utilize the gradient descent (GD) method \cite{gradient_descent_method} to find the parameters that minimize \(d\) on a fixed and finite duration in time \([t_0, t_f]\) such that \(t_0 < t_f\). Notably, the GD is applied on the model's parameters space such that the gradient for each configuration of parameter values is numerically obtained using the five-point stencil numerical scheme \cite{numerical_scheme}. The result of this process is the model's parameters' values that result in the closest clinical outcome of the given train set, divided into the socio-economic groups presented in the data. This step is shown to provide efficient initial fitting of the model's parameters \cite{liza_latest}. Nonetheless, due to the small amount of data, it is probably not spread across the domain properly which highly limits the usefulness of this method \cite{gradient_descent}. To this end, for the second step, we randomly generate a new sample such that the features' values are range between the minimum and maximum of the values in the train set and chosen in a uniform distribution. Once a sample is generated, we test if the model's prediction of the clinical outcome's error is less or higher than the one obtained from a \(k\) nearest neighbor algorithm, on average, for the optimal choice of \(k\). If the sample does not fulfill this condition, it is removed. Otherwise, it is added to the train set. This process repeats itself until \(n \in \mathbb{N}\) samples are added. Intuitively, this step allows to generate a \say{filling} of synthetic samples based on the first step. Hence, afterward, the first step is repeated for the next train with a synthetic set. Using paired one-tail T-test, we check if the latter provides statistically significant better results on the test set. If it does not, we repeat the previous step. Otherwise, the train, synthetic, and test sets are merged and the last step is taking place. Namely, at this point, using the synthetic data from the second step, we remedy the shortcoming of the first step. Nonetheless, as we used the \(k\) nearest neighbor algorithm in the second step which is known to poorly extrapolate \cite{knn_bad_1,knn_bad_2}. Further improvement of the fitting robustness can be accomplished by utilizing other machine learning models for better extrapolation over the socio-demographic parameter space. Thus, for the third step, we use the Tree-based Pipeline Optimization Tool (TPOT) \cite{tpot} automatic machine learning framework to search a large number of machine learning pipelines. For each pipeline, we test its performance on the validation set. The machine learning pipeline with the best performance, given a pre-defined metric \((M)\), is chosen. This process is repeated for each \(i \in [1, \dots, k]\) fold such that the overall prediction is the average of each machine learning model achieved at each fold. Overall, Fig. \ref{fig:fit} provides a schematic view of the proposed fitting method and summarizes its main steps. 

\section{Theoretical Analysis}
\label{sec:theory}

Here, we theoretically analyze the proposed model (Eq. \ref{eq:1}--\ref{eq:6}). We start by proving that the model always has a unique solution. Then, we identify the model's equilibria and analyze their stability.

\subsection{Solution existence and uniqueness}
In order to show that the proposed model has a solution and it is unique, we utilize the Picard–Lindelöf theorem \cite{appendix}. Formally, the Picard–Lindelöf theorem states that if \(D \subset \mathbb{R} \times \mathbb{R}^n\) is a closed rectangle with \((t_0, y_0) \in D\) and \(f: D \rightarrow \mathbb{R}^n\) is a function that is continuous in \(t\) and Lipschitz continuous in \(y\); then there exists some \(\epsilon > 0\) such that the initial value problem:
\begin{equation}
    y'(t) = f(t, y(t)), y(t_0) = y_0,
\end{equation}
has a unique solution \(y(t)\) on the interval \([t_0 - \epsilon, t_0 + \epsilon]\). Thus, for our case \(y(t) := (B(t), E(t), T_i(t), T_u(t), H_u(t), H_i(t))\). In order to use the Picard–Lindelöf theorem, we first need to show that Eqs. (\ref{eq:1}-\ref{eq:6}) is continuous in \(t\) and Lipschitz continuous in \(y\). To this end, let us consider a finite duration in time \([0, T]\) such that \(T < \infty\). Next, the interaction between the of the unknown solution, \(y\), has terms of a linear form and of the form \(y_i y_j\), the function \(f\) such that \(dy(t)/dt = f(t, y(t)\) is \(C^1\) which implies that it also locally satisfies Lipschitz condition and continuous in \(t\) \cite{sveta_appendix}. Therefore, one can apply the Cauchy–Lipschitz theorem  \cite{final_appendix} which leads to the result of the existence and uniqueness of the solution to
Eq. (\ref{eq:1}--\ref{eq:6}), on any finite interval \([0,T]\).

Hence, we show that a solution exists. As such, we need to show that for any non-negative initial condition, the solution is non-negative. To this end, let us assume a non-negative initial condition \((B(0) \geq 0, E(0) \geq 0, T_i(0) \geq 0, T_u(0) \geq 0, H_i(0) \geq 0, H_u(0) \geq 0)\) for the proposed model (Eq. (\ref{eq:1}-\ref{eq:6})). Now, let us focus on the fourth equation. By dividing by \(T_u(t)\), one obtains:
\begin{equation}
    T_u'(t)/T_u(t) = \lambda - p_2(t)B(t) - p_3(t)E(t).
\end{equation}
Computing the integral for \(t\), we obtain that:
\begin{equation}
    T_u(t) = e^{\int(\lambda - p_2(t)B(t) - p_3(t)E(t))dt}T_u(0),
\end{equation}
Since \(T_u(0) \geq 0\), we obtain \(T_u(t) \geq 0\) for any value of \(t\). For the fifth equation, we have a second-order equation in \(H_u(t)\). Namely, for \(H_u(t)\) the fifth equation is a Riccati equation, and therefore the solution of the equations takes the form:
\begin{equation}
    H_u(t) = -\frac{\int \big ( a(t) + \frac{db(t)/dt}{b(t)} \big ) dt}{b(t)}
\end{equation}
where \(a(t) = p_7(t) - p_8(t)B(t) - \frac{p_7(t) \big (  H_i(t) + T_u(t) + T_i(t)\big )}{H_m}, b(t) = -p_7(t)/H_m\). It is easy to see that as long as \(0 \leq p(t)\) than \(H_u(t) \geq 0\) for any value of \(t\). Using the fact that \(H_u(t) > 0\), we use the same method utilized for the third equation and obtain that \(0 \leq H_i(t) \) for any \(t\). The first and second equations yield that \(0 \leq E(t)\) and \(0 \leq B(t)\) for any \(t\) since \(T_i(t), H_i(t), T_u(t)\), and \(H_u(t)\) are non-negative. Thus, we show that the proposed model's solution is non-negative for any value of \(t > 0\) if the initial condition is non-negative. 

\subsection{Equlibria and stability analysis}
In order to better understand the bio-mathematical properties of the proposed model, we computed the equilibria states of the proposed model and their stability properties. Recall that an equilibrium state is reached when the system does not change without outside intervention. As such, to compute the equilibria states of the systems, we set the left side of the equations in Eqs. (\ref{eq:1}-\ref{eq:6}) to zero and solve for the vector \([B(t), E(t), T_i(t), T_u(t), H_u(t), H_i(t)]\). Following this, one obtains two equilibria states: \(B(t) = \sum_{m=0}^{N-1}b\delta(t-m\tau) / \mu_B, E(t) = T_i(t) = T_u(t) = H_u(t) = H_i(t) = 0\) and \(B(t) = E(t) = T_i(t) = T_u(t) = H_i(t) = 0, H_u(t) = H_m\). The first one is not mathematically valid, as for different values of \(t\), \(B(t)\) would have different values during \(0 \leq t \leq (N-1)\tau\). As such, this equilibrium is well-defined for \(t > (N-1)\tau\) which results in \(B(t) = E(t) = T_i(t) = T_u(t) = H_i(t) = H_u(t) = 0\) which is the trivial equilibrium where no dynamic takes place. For the second case, it is also trivial in the sense that all cells are healthy. 

In order to obtain the equilibria states' stability of the two equilibria states, we first compute the Jacobian matrix of Eq. (\ref{eq:summary}), following Routh–Hurwitz stability criterion \cite{stability}: 
\begin{equation}
    \begin{array}{l}
         J = \begin{pmatrix}
         -p_1E-p_2T_u-\mu_B & -p_1B & 0 & -p_2B & -p_8B & 0 \\
         p_4E & -\mu_E+p_4E-p_5T_i-p_6H_i & \alpha-p_5E & 0 & 0 & \alpha-p_6E \\
         0 & -p_3T_i & -p_3E & p_2B & 0 & 0 \\
         -p_2T_u & -p_3T_u & 0 & \lambda-p_2B-p_3E & 0 & 0 \\
         -p_8H_u & 0 & -1/H_m & -1/H_m & p_7-p_8B-2H_u/H_m & -1/H_m \\
         p_8H_u & -p_9E & 0 & 0 & p_8B & -p_9E \\
\end{pmatrix}
    \end{array}.
\end{equation}
Now, following the Hartman–Grobman theorem \cite{stability_2}, by setting each equilibrium state to \(J\) we obtain: 
\begin{equation}
    \begin{array}{l}
         J_{trivial} = \begin{pmatrix}
         -\mu_B & 0 & 0 & 0 & 0 & 0 \\
         0 & -\mu_E & \alpha & 0 & 0 & \alpha \\
         0 & 0 & 0 & 0 & 0 & 0 \\
         0 & 0 & 0 & \lambda & 0 & 0 \\
         0 & 0 & -1/H_m & -1/H_m & p_7 & -1/H_m \\
         0 & 0 & 0 & 0 & 0 & 0 \\
\end{pmatrix}
    \end{array},
\label{eq:j_1}
\end{equation}
and 
\begin{equation}
    \begin{array}{l}
         J_{healthy} = \begin{pmatrix}
         -\mu_B & 0 & 0 & 0 & 0 & 0 \\
         0 & -\mu_E & \alpha & 0 & 0 & \alpha \\
         0 & 0 & 0 & 0 & 0 & 0 \\
         0 & 0 & 0 & \lambda & 0 & 0 \\
         -p_8H_m & 0 & -1/H_m & -1/H_m & p_7+2 & -1/H_m \\
         p_8H_m & 0 & 0 & 0 & 0 & 0 \\
\end{pmatrix}
\label{eq:j_2}
    \end{array}.
\end{equation}
We compute the eigenvalues of Eqs. (\ref{eq:j_1}-\ref{eq:j_2}), obtaining that both have at least one eigenvalue that equals zero due to the third line being full of zeros, respectively. Hence, both equilibria are unstable. 

\section{Empirical Analysis}
\label{sec:results}

\subsection{Data acquisition and preprocessing}
We obtained retrospective data from the Bnai-Zion Medical Center\footnote{We refer the interested reader to \url{https://www.gov.il/he/departments/b-zion-health-center/govil-landing-page} (In Hebrew)} (Israel) from 2008 and 2017 \cite{different_b_model}. The data was anonymously extracted from the hospital's records under the following restrictions: 1) patients are adults (\(>18\) years old); 2) Patients received the standard BCG-based treatment for their non-invasive bladder cancer; 3) Patients were admitted between 2008 and 2017, in which period where all patients obtained the same treatment protocol for a non-invasive BC. In total, 417 BC patients are included, representing the entire patient population satisfying these three conditions. For each patient, we extracted the size of the cancer tumor at the beginning and end of the treatment alongside the four socio-demographic properties outlined above (gender, age, smocking, and weight). Additionally, for each patient, we extract the amount of BCG injected \((b)\), the number of BCG injections \((N)\), the duration between every two consecutive BCG injections \((\tau)\), BCG's decaying rate \((\mu_B)\), and the number of cells in the bladder \((H_m)\). The BCG-treatment-related parameters (\(b, N, \tau\)) are defined by the standard treatment protocol \cite{MoralesEidingerBruce} to be \(2.8 \cdot 10^8, 6,\) and \(7\) days. The amount of cells in the bladder is highly linear to the patient's age and weight \cite{volume_1,volume_2}. As such, we use the formula proposed in \cite{volume_3} to approximate this value. The above is summarized in Table \ref{table:parameter_values}. 

In addition, as \(T_i(0) + T_u(0)\) and \(T_i(t_f) + T_u(t_f)\) that required by the fitting procedure are not measurable \say{as is} rather than the polyp's volume it measured, one is required to map between these two values. To this end, we used the average volume of bladder cancer cells \cite{cancer_cell_volume} and the polyp's volume to approximate \(T_i(0) + T_u(0)\) and \(T_i(t_f) + T_u(t_f)\).

\begin{table}[!ht]
    \centering
    \begin{tabular}{p{0.15\textwidth}p{0.5\textwidth}p{0.15\textwidth}}
    \hline
    \textbf{Parameter} & \textbf{Description} & \textbf{Average value} \\ \hline
     \(\mu_B\) &	BCG half-life in hours  [\(t^{-1}\)] & \(4.16 \cdot 10^{-3}\)  \\
     \(b\) & Dose of BCG & \(2.8 \cdot 10^{6}\) \\
     \(N\) & Number of BCG injections [\(1\)]  & \(6 \cdot 10^0\) \\
     \(\tau\) & Duration in hours between two consecutive BCG injections [\(t\)]  & \(1.68 \cdot 10^{2}\) \\
     \(H_m\) & The number of healthy cells in the bladder without cancer  as a function of the patient's gender and weight [\(1\)]  & \(1.84 \cdot 10^9\) \\ \hline
    \end{tabular}
    \caption{The model's parameter definitions and average values as adopted from \cite{math_good_1,teddy_cells}.}
    \label{table:parameter_values}
\end{table}

\subsection{Parameter fitting}
\begin{figure}
    \centering
    \includegraphics[width=0.75\textwidth]{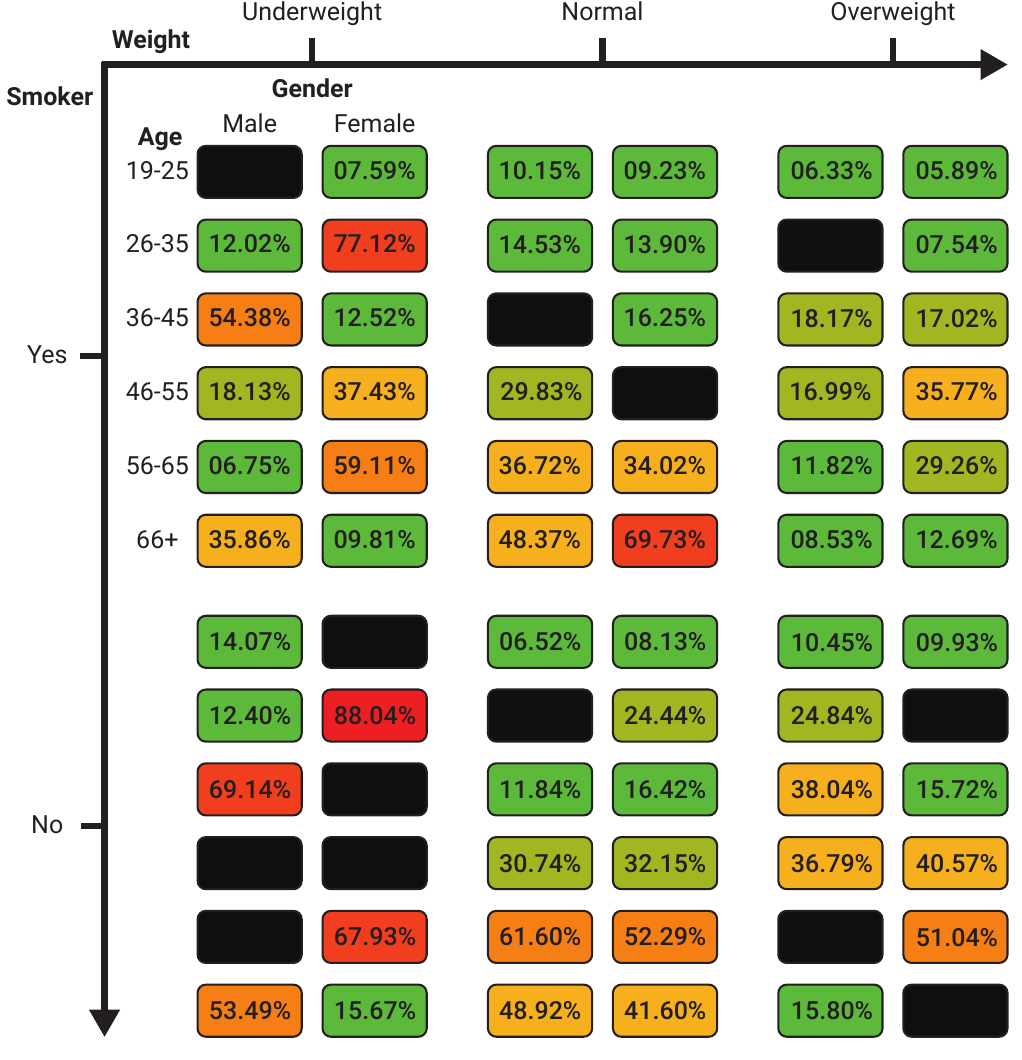}
    \caption{The average over \(k=5\) folds relative absolute error of the model's fitting on the test set. The black boxes indicate that no patients belong to this combination of socio-demographic values.}
    \label{fig:fitting_results}
\end{figure}

In order to use the proposed parameter fitting procedure, one is required to define a fitting metric. Hence, to make sure the model is able to better predict the outcome of a treatment procedure given an initial condition, we used the relative absolute error metric between the model's prediction and the historical data. Formally, let us denote the cancer population the model's prediction, and historical using \(T_{t_f}^m\) and \(T_{t_f}^h\) respectively. The relative mean absolute error than takes the form \(RMAE := \frac{1}{|P|}\sum_{p \in P} |T_{t_f}^m - T_{t_f}^h|/(T_{t_f}^h)\) such that \(P\) is the population of patients. Now, following the proposed fitting procedure (see Section \ref{sec:fitting}) and the acquired data, we fitted the model and obtained \(28.51 \pm 3.76\) relative absolute error on the test set of \(k=5\) folds. Fig. \ref{fig:fitting_results} shows the distribution of the relative mean absolute error among the four socio-demographic features - smoking habits, weight group, age group, and gender. We use black boxes to indicate configurations that could not be assessed given the available data (i.e., no single test case was available). Notably, non-smokers with underweight have the worst fitting results compared to other configurations. This may be partially attributed to the small number of samples within these groups during the training phase - 13 in total. In addition, we compared the performance of the model's fitting, divided into smokers and non-smokers groups, using a two-tailed T-test \cite{t_test}. We obtained that zero is not included in the confidence interval of the T-test, which indicates that the two groups are not statistically significantly different. In a similar manner, when comparing the model's fitting performance between male to female patients using a T-test, we obtain \(p = 0.061\) which indicates no statistically significant difference. Moreover, when comparing the weight categories using an ANOVA test with post-hoc Tukey pairwise tests \cite{anova}, we see that the normal weighted category obtains statistically worse results compared to the two other categories \(p < 0.05\). Similarly, the 56-65 and 65+ age groups obtained statistically significantly worse results compared to the younger age groups with \(p < 0.05\). 

\subsection{Evaluation}

\begin{figure}[!ht]
    \centering
    \includegraphics[width=0.99\textwidth]{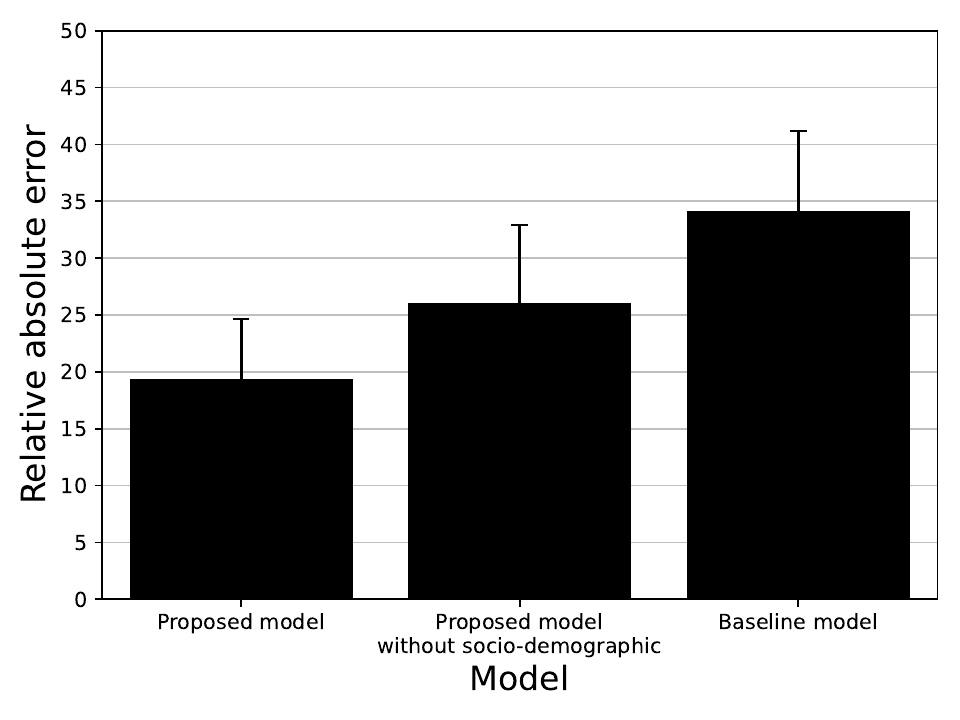}
    \caption{A comparison between the proposed model, the proposed model without socio-demographics, and the standard model \cite{BunimovichShochat}.}
    \label{fig:compare}
\end{figure}

We evaluate the proposed model by comparing it to two baselines: First, we use a narrow version of the proposed model which \textit{does not use of the socio-demographics} at all. Second, we use the original model proposed by \cite{BunimovichShochat} as provided by the authors. Fig. \ref{fig:compare} shows the relative mean absolute error of the three models for 50 samples. We choose the samples such that each sample would have a unique combination of the socio-demographic values. One can clearly see that the proposed model outperforms both other baselines with an average relative absolute error of \(19.38 \pm 5.27\) compared to \(26.04 \pm 6.85\) and \(34.18 \pm 7.02\), respectively. 

\section{Discussion}
\label{sec:discussion}
In this study, we presented a mathematical model to describe the dynamics of BCG-based immunotherapy for BC which incorporates socio-demographic parameters, such as age, gender, smoking behavior, and weight, to capture the heterogeneity in the patient populations and provide a more accurate treatment outcome prediction. By categorizing patients into distinct, easily identifiable, socio-demographic groups, we acknowledge the diverse characteristics of BC patients and their potential influence on the treatment outcome \cite{bc_summary_1,bc_summary_2,bc_summary_3}. This approach allows one to tailor more personalized models and thus obtain a better understanding of how different patient profiles may respond to BCG-based immunotherapy. To this end, we used a novel fitting procedure that uses machine learning methods with only partial and sparse historical data to fit the personalized model for each socio-demographic group. 

Our results strongly suggest that the currently unexploited socio-demographics may encompass clinically relevant information which underlines the BCG-based immunotherapy treatment of non-invasive BC. Most notably, the proposed model which takes these socio-demographics into account outperforms the baseline model proposed in \cite{BunimovichShochat} by \(14.8.\%\) as well as the same model which does not consider socio-demographics by \(8.14\%\), as shown in Fig. \ref{fig:compare}. Arguably, these improvements should be attributed to the more personalized, thus accurate, ODE model. We observe further support for this observation in Fig. \ref{fig:fitting_results} which shows that different socio-demographic groups are associated with different fitting performances. This outcome can be associated with two factors: First, the differences in the amount of available data used to fit the model. Second, the clinical processes and dynamics that change as a result of belonging to each socio-demographic group. The effect of the first factor is clearly revealed for the non-smoker and the underweight group as a low amount of fitting data results in the highest, on average, fitting error. In a complementary manner, the second factor can be seen in older patients having higher fitting errors, on average, compared to younger patients while being a large portion of the dataset. 

It is important to note that our model and results have certain limitations. First, the proposed model does not take into consideration the geometrical configuration of the bladder and therefore the spatial dynamics of the BCG treatment which has shown to have a critical role in the treatment outcomes and optimal treatment protocol for patients \cite{teddy_cells,newPaper,oldPaper}. As such, one can introduce these extensions to obtain a more realistic model. Second, the model assumes that a patient's socio-demographic properties remain constant during the course of BCG treatment. While this assumption is accepted due to the short duration of the treatment (i.e., around two months), it may not fully capture potential changes in patient characteristics that could influence treatment response. Third, the proposed model is partially personalized. Potentially, by using more parameters, one can obtain better personalization and even further improve the model's prediction accuracy \cite{ariel_final}. Finally, our model focuses on the treatment protocol clinical outcome prediction rather than suggesting an optimal personalized treatment as other studies do \cite{math_good_1}. In future work, we plan to further investigate this direction. 

Overall, the proposed model provides a low (to no) overhead method for clinicians while providing a statistically significant improvement in the prediction of treatment outcomes. Thus, it can be easily adopted by healthcare professionals and help provide better treatment, saving and improving many lives. In the same manner, our proposed fitting method can be utilized in a broad spectrum of models as it assumes as little as possible on the historical data and nothing on the model itself.  

\section*{Declarations}
\subsection*{Funding}
This research did not receive any specific grant from funding agencies in the public, commercial, or not-for-profit sectors.

\subsection*{Conflicts of interest/Competing interests}
None.

\subsection*{Data availability}
Due to the sensitivity of the data, it is available from the authors upon reasonable request. 

\subsection*{Acknowledgement}
The authors wish to thank Sarel Halachmi for providing the data used in this work. Elizaveta Savchenko wishes to thank Ariel University's financial support during this research.  
 
\bibliography{biblio}
\bibliographystyle{unsrt}

\end{document}